\pdfoutput=1
\documentclass{article}
\usepackage[T1]{fontenc}
\usepackage{arxiv_preprint,times}
\usepackage{amsmath,amssymb,booktabs,multirow,colortbl,xcolor,pifont,makecell,graphicx}
\usepackage{microtype,subcaption,placeins,hyperref,url}
\hypersetup{hidelinks,pdfauthor={Bing Fan, Anwesa Choudhuri, Zhongpai Gao, Meng Zheng, Benjamin Planche, Terrence Chen, Heng Fan, Ziyan Wu},pdftitle={STEP: Surgical Point Tracking with Curated Trajectories and Physical Endpoints}}
\graphicspath{{fig/}}
\definecolor{mygray}{gray}{0.94}
\definecolor{highlight}{RGB}{238,246,239}
\newcommand{\improve}[1]{{\scriptsize\textcolor[RGB]{45,112,76}{~(#1)}}}

\title{STEP: Surgical Point Tracking from \\ Sparse Endpoints and Pseudo-trajectories}
\author{%
Bing Fan$^{\mathbf{1,2,*}}$, Anwesa Choudhuri$^{\mathbf{1,\dagger}}$, Zhongpai Gao$^{\mathbf{1,\dagger}}$, Meng Zheng$^{\mathbf{1}}$, \\
\bfseries Benjamin Planche$^{\mathbf{1}}$, Terrence Chen$^{\mathbf{1}}$, Heng Fan$^{\mathbf{2}}$, Ziyan Wu$^{\mathbf{1}}$ \\[3pt]
\normalfont $^{1}$United Imaging Intelligence, Boston, MA, USA \\
\normalfont $^{2}$University of North Texas, Denton, TX, USA
}
\begin{document}
\raggedbottom
\maketitle
\thispagestyle{fancy}
{\renewcommand{\thefootnote}{}\footnotetext{$^{*}$The work conducted by the first author was carried out during an internship at United Imaging Intelligence.}\footnotetext{$^{\dagger}$Corresponding authors.}}

\begin{figure}[!ht]
\centering
\includegraphics[width=\linewidth]{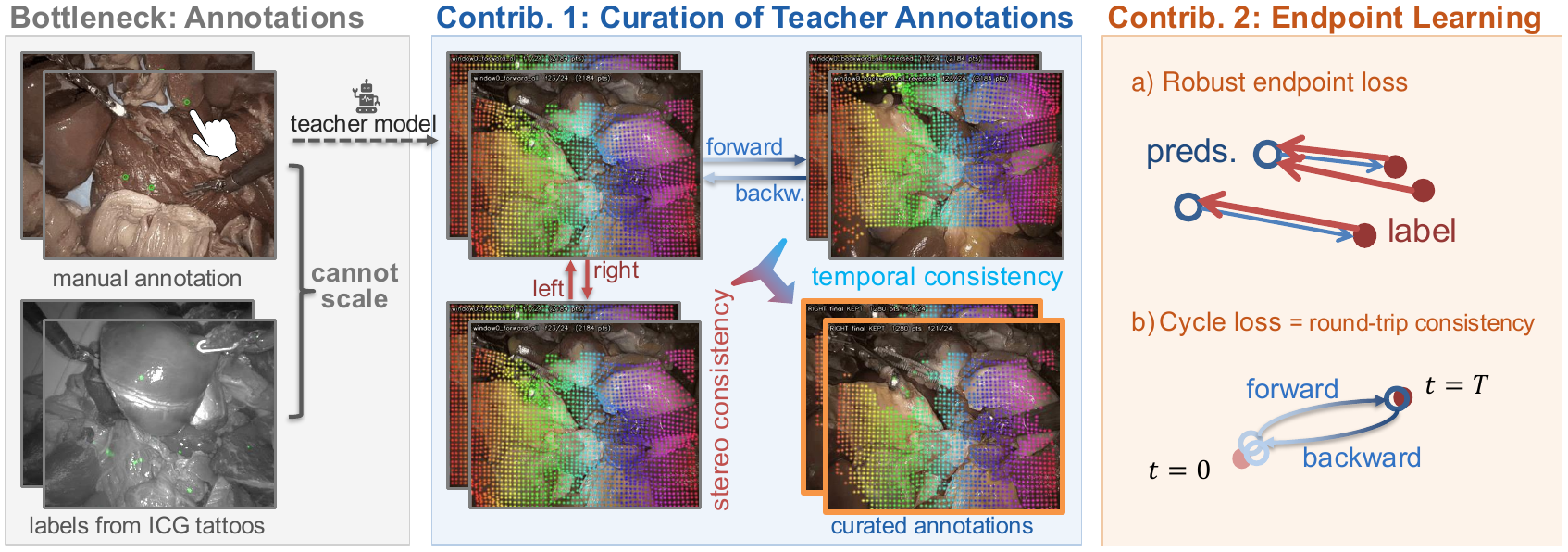}
\caption{\textbf{Scaling supervision for surgical point tracking.} Manual trajectories require repeated annotation. STEP combines curated trajectories and existing physical endpoints to train a causal surgical tracker.}
\label{fig:teaser}\label{fig:paradigm}
\par\vspace{0.4em}\hrule height 0.4pt
\end{figure}
\FloatBarrier

\begin{abstract}
Surgical point tracking requires persistent correspondence through tissue deformation
and occlusion, yet dense trajectory annotations are costly to obtain. We present
STEP, a two-stage framework that learns from surgical video and sparse physical
endpoints. First, a frozen offline teacher generates candidate trajectories, and
temporal reversibility and stereo consistency select dense supervision for a causal
student. Second, a robust physical endpoint loss anchors terminal correspondence, while
a round-trip loss constrains temporal consistency.
A gradient-free prefix exposes training to accumulated drift, while a final
differentiable window bounds activation memory. The trained tracker processes a
single RGB stream with fixed weights. In STIR-to-VL-SurgPT transfer, temporal--stereo
union selection reduces endpoint error by 7.0\% relative to unfiltered trajectories.
On a post-hoc diagnostic of 29 VL-SurgPT clips excluded from the endpoint-training
manifest, STEP reduces endpoint error from 31.52 to 16.80 pixels relative to
dense-label fine-tuning. These results support combining curated trajectories with
physical observations to train surgical point trackers.

\end{abstract}

\section{Introduction}
Tracking tissue points through surgical video provides correspondence for motion
estimation, deformation modeling, registration, and robotic perception. Tracking
becomes difficult when tissue deforms under tool interaction, specular reflections
change its appearance, and instruments hide the target. General-purpose trackers
learn strong correspondence priors from synthetic and real-world video
\citep{doersch2023tapir,karaev2025cotracker3,aydemir2026trackonr}; adapting these
priors to surgery requires supervision that captures these conditions.

Surgical video offers two complementary sources of supervision. Unlabeled footage
captures diverse motion and appearance, but obtaining dense manual trajectories is
expensive, especially through deformation and occlusion. Physical markers offer a
second resource: STIR and VL-SurgPT record tissue-point positions at the beginning
and end of a clip using fluorescent observations
\citep{schmidt2024stir,zhou2026vlsurgpt}. These measurements establish correspondence
without tracing every frame. However, they leave the intervening path unobserved,
and marker localization, matching, and switching between fluorescence and
white-light imaging can introduce uncertainty. A tracker must learn dense correspondence from video and anchor predictions
to sparse measurements.

These signals address different ambiguities. A teacher can produce smooth,
reversible tracks that gradually attach to the wrong tissue region, while an
accurate endpoint does not specify the route taken through an occlusion. Training
only on short windows also avoids the accumulated errors encountered over a
complete surgical clip. This motivates combining dense video-derived targets
with measured terminal correspondence, and applying endpoint supervision to the
student's predictions.

We introduce STEP (Surgical Tracking from Sparse Endpoints and Pseudo-trajectories),
which combines these resources in two training stages (Figure~\ref{fig:teaser}).
The first stage curates dense trajectories from a frozen offline teacher.
Temporal reversibility checks whether forward tracking returns to the query under
reverse tracking; stereo consistency checks agreement with binocular geometry.
The tests accept complementary candidates, so we retain their union on stereo
video and use temporal selection on monocular video. The resulting trajectories
and visibility masks supervise a causal student throughout each training window.

The second stage uses physical endpoints to refine long-range correspondence.
The endpoint loss uses known point identities when available and compares
unordered marker sets through robust nearest-neighbor distances otherwise.
A round-trip loss adds temporal supervision by requiring tracks to return to their
starting locations. To train on long clips with bounded activation memory, the
student processes earlier frames without gradients and initializes a final
training window from its own predicted positions. Training thus encounters accumulated streaming errors.

Dense teacher trajectories, measured endpoints, and round trips play different
roles: they supply local correspondence, anchor terminal positions, and constrain
temporal consistency. Both stages update weights before deployment; inference
then requires only initial queries and a single RGB video stream. Our main
contributions are:
\begin{itemize}
\setlength{\itemsep}{2pt}\setlength{\parsep}{0pt}
\item \textbf{Surgical trajectory curation.} Temporal and stereo checks select
visibility-aware pseudo-trajectories from surgical video.
\item \textbf{Endpoint-guided streaming tracking.} Physical endpoint and round-trip
objectives train on accumulated prediction error within a bounded gradient window.
\item \textbf{Empirical findings.} Union selection lowers transfer EPE by 7.0\%.
On 29 diagnostic clips, STEP lowers Track-On-R EPE by 46.7\% versus dense-label
fine-tuning.
\end{itemize}

\section{Related Work}
\textbf{General-purpose and surgical tracking.}
Tracking Any Point estimates the trajectory and visibility of arbitrary queries,
with TAP-Vid providing a common evaluation protocol~\citep{doersch2022tapvid}.
TAPIR refines point correspondences, CoTracker3 reasons jointly over tracks,
and MFT composes dense flow over long horizons
\citep{doersch2023tapir,karaev2025cotracker3,neoral2024mft}.
Track-On and Track-On-R carry memory across frames for online tracking
\citep{aydemir2025trackon,aydemir2026trackonr}. Surgical methods incorporate
stereo geometry and deformation (SENDD), adaptive templates (Ada-Tracker), or
dense surgical annotations (SurgMotion)
\citep{schmidt2023sendd,guo2024adatracker,zhan2025surgmotion};
\citet{schmidt2024review} review this setting. STEP focuses on the supervision
needed to adapt pretrained trackers to surgical correspondence.
In particular, an offline teacher can use later frames to resolve earlier
ambiguities, while a streaming tracker must update its estimate as each frame
arrives. STEP transfers the former's predictions into training targets for the
latter. Endpoint refinement trains on the student's causal predictions, aligning
supervision with streaming inference.

\textbf{Learning from video and physical observations.}
BootsTAP and CoTracker3 use predictions on unlabeled video to expand training
supervision, while Track-On-R uses a learned verifier to select reliable
pseudo-labels~\citep{doersch2024bootstap,karaev2025cotracker3,aydemir2026trackonr}.
In surgery, STIR and VL-SurgPT provide physically measured correspondence from
tissue markings~\citep{schmidt2024stir,zhou2026vlsurgpt}. STEP connects these
resources: consistency-selected teacher trajectories provide dense surgical
supervision, and physical endpoints refine terminal correspondence. It uses existing benchmark measurements without collecting new manual trajectories.

\textbf{Geometric and temporal consistency.}
Forward--backward agreement identifies reliable correspondences in unsupervised
flow~\citep{meister2018unflow}, while temporal cycles support correspondence
learning in TimeCycle and contrastive random walks
\citep{wang2019timecycle,jabri2020crw}. Stereo geometry constrains matches across
views~\citep{hartley2004mvg}, with left--right consistency used in depth learning
and temporally consistent disparity estimated by Stereo Any Video
\citep{godard2017stereo,jing2025stereoanyvideo}. STEP uses temporal and stereo
consistency to select training targets, then combines round-trip supervision
with physical endpoints during streaming training.

\section{Curated Surgical Supervision}
\label{sec:data}\label{sec:stage1}

Our curation pipeline converts surgical video into trajectories with per-frame
visibility and validity masks. It supports monocular and stereo inputs from
STIR, VL-SurgPT, StereoMIS, Hamlyn, and SurgMotion
\citep{schmidt2024stir,zhou2026vlsurgpt,hayoz2023stereomis,hamlyn,zhan2025surgmotion}.
The experiments use STIR and VL-SurgPT to study selection and endpoint-guided
training; Appendix~\ref{app:data} describes the source inventories.

\subsection{Candidate Trajectories and Consistency Selection}
\label{sec:selection}

\textbf{Offline candidate generation.}
A frozen AllTracker teacher~\citep{harley2025alltracker} predicts trajectories
from regular point grids over windows of at most 750 frames. An independently
seeded reverse pass supplies correspondences for temporal checking. We retain
coordinates and per-frame visibility for each candidate. Visibility is defined
by $1-\mathrm{visibility}\cdot\mathrm{confidence}\leq0.4$ using the teacher's
outputs. The teacher accesses the full window to supervise a causal student.

\begin{figure}[t]
    \centering
    \includegraphics[width=\textwidth]{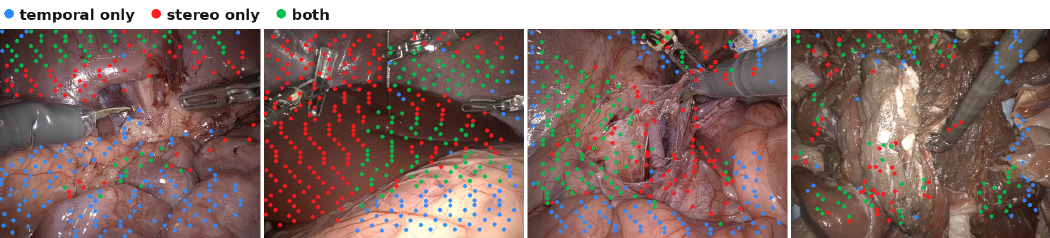}
    \caption{\textbf{Temporal and stereo consistency selection.} STIR candidates accepted by temporal checks only (blue), stereo checks only (red), or both (green). Stage~1 retains their union on stereo sequences.}
    \label{fig:selection}
\end{figure}

\textbf{Temporal consistency selection.}
We first assess whether a candidate remains consistent under temporal reversal,
following the forward--backward principle commonly used in unsupervised flow and
correspondence learning~\citep{meister2018unflow,wang2019timecycle}. For a query
$q_i$ at frame $t_a$, let $F_{a\rightarrow b}(q_i)$ denote its predicted location
at frame $t_b$. An independently seeded reverse pass is interpolated at this
location to obtain the corresponding return position. We define the temporal
round-trip error as
\begin{equation}
e_T(i)=
\left\|
B_{b\rightarrow a}\!\left(F_{a\rightarrow b}(q_i)\right)-q_i
\right\|_2.
\label{eq:temporal}
\end{equation}
A candidate is included in $\mathcal{T}_T$ if $e_T(i)<\tau_T$ and the
correspondence is visible in both passes. We use $\tau_T=3$ pixels in the
teacher coordinate system. Per-frame visibility is retained for selected
trajectories, so frames in which the target is occluded do not introduce
artificial positional supervision.

\textbf{Stereo consistency selection.}
For stereo sequences, we additionally evaluate candidates using cross-view
geometry. Calibrated left--right correspondence~\citep{hartley2004mvg} is
estimated using Stereo Any Video~\citep{jing2025stereoanyvideo}. Let $d_i(t)$
denote the disparity-induced offset predicted at candidate $i$ and time $t$, and
let $s_i(t)$ denote the offset between its associated trajectories in the two
views under the same coordinate convention. Their disagreement is
\begin{equation}
e_S(i,t)=\left\|d_i(t)-s_i(t)\right\|_2.
\label{eq:stereo}
\end{equation}
A stereo correspondence is consistent when $e_S(i,t)<\tau_S$. We associate cross-view tracks within 3 pixels and require disparity agreement within $\tau_S=8$ pixels over visible frames.

Temporal and stereo consistency provide complementary evidence rather than
identical filters. As illustrated in Fig.~\ref{fig:selection}, some candidates
satisfy only the temporal criterion (blue), others only the stereo criterion
(red), while a subset satisfies both (green). We therefore retain the union
\begin{equation}
\widetilde{\mathcal{T}}
=
\mathcal{T}_T \cup \mathcal{T}_S
\end{equation}
for stereo sequences, rather than requiring their intersection. For monocular
sequences, or windows without valid stereo estimates, we use
$\widetilde{\mathcal{T}}=\mathcal{T}_T$. The retained per-frame visibility masks
determine which point--frame pairs contribute supervision during training.

\textbf{Training with curated trajectories.}
The selected trajectories provide pseudo-supervision for Stage~1 (S1). Starting
from a released checkpoint, we train the causal student using teacher coordinates
$\widetilde{p}_i^t$ only at valid point--frame pairs indicated by $w_i^t$. The
positional component of the objective is
\begin{equation}
\mathcal{L}_{\mathrm{pseudo}}
=
\frac{
\sum_{i,t} w_i^t\,
\ell_{\mathrm{track}}
\!\left(\widehat{p}_i^t,\widetilde{p}_i^t\right)
}{
\sum_{i,t} w_i^t
},
\label{eq:pseudo}
\end{equation}
where $\ell_{\mathrm{track}}$ is the student's native tracking loss. This stage
produces initialization $\theta_1$ for endpoint-guided refinement.
Section~\ref{sec:curation-results} evaluates selection through cross-dataset transfer.

The temporal test used here selects fixed teacher targets before optimization.
The round-trip objective in Section~\ref{sec:stage2} instead acts on the student's
current predictions and updates its parameters. This distinction lets curation
exploit an offline teacher's temporal context while endpoint-guided training
improves the student that will run causally at deployment.

\subsection{Supervision Coverage and Visibility}
Each clip-view record stores frame indices, image size, coordinates, per-frame
validity, and temporal and stereo acceptance masks. During training, frames and
coordinates are resized together; non-finite, out-of-frame, and teacher-occluded
positions are excluded from the positional loss. Thus, accepting a trajectory
does not imply supervising every point along it.

The generated inventories contain 493,893 candidate trajectories for VL-SurgPT
and 1,685,211 for STIR, spanning approximately 103M and 300M point-frames,
respectively (Table~\ref{tab:data}). These inventories precede training selection
and split filtering. Each trajectory carries its own validity mask, so increasing
coverage does not require treating uncertain or occluded positions as observed
coordinates. Together, the source records and masks allow the same candidate pool
to support different selection policies without regenerating teacher predictions.

For VL-SurgPT Stage~1, excluding the 150 evaluation clips leaves 604 source clips,
of which 596 yield usable windows. STIR selection experiments share 1,726
clip-view windows from 566 stereo clips and use the same optimization budget.
Source-specific preprocessing preserves the correspondence between image content
and trajectory coordinates. StereoMIS tracks follow the decoded packet timeline,
while Hamlyn uses temporal selection and a first-frame grid check to remove
static border candidates. Such regions can appear temporally consistent despite
providing no tissue-motion supervision. Appendix~\ref{app:data} details source
inventories and preprocessing.

\section{Endpoint-Guided Streaming Tracking}
\label{sec:method}\label{sec:stage2}

\begin{figure}[t]
\centering
\includegraphics[width=\linewidth]{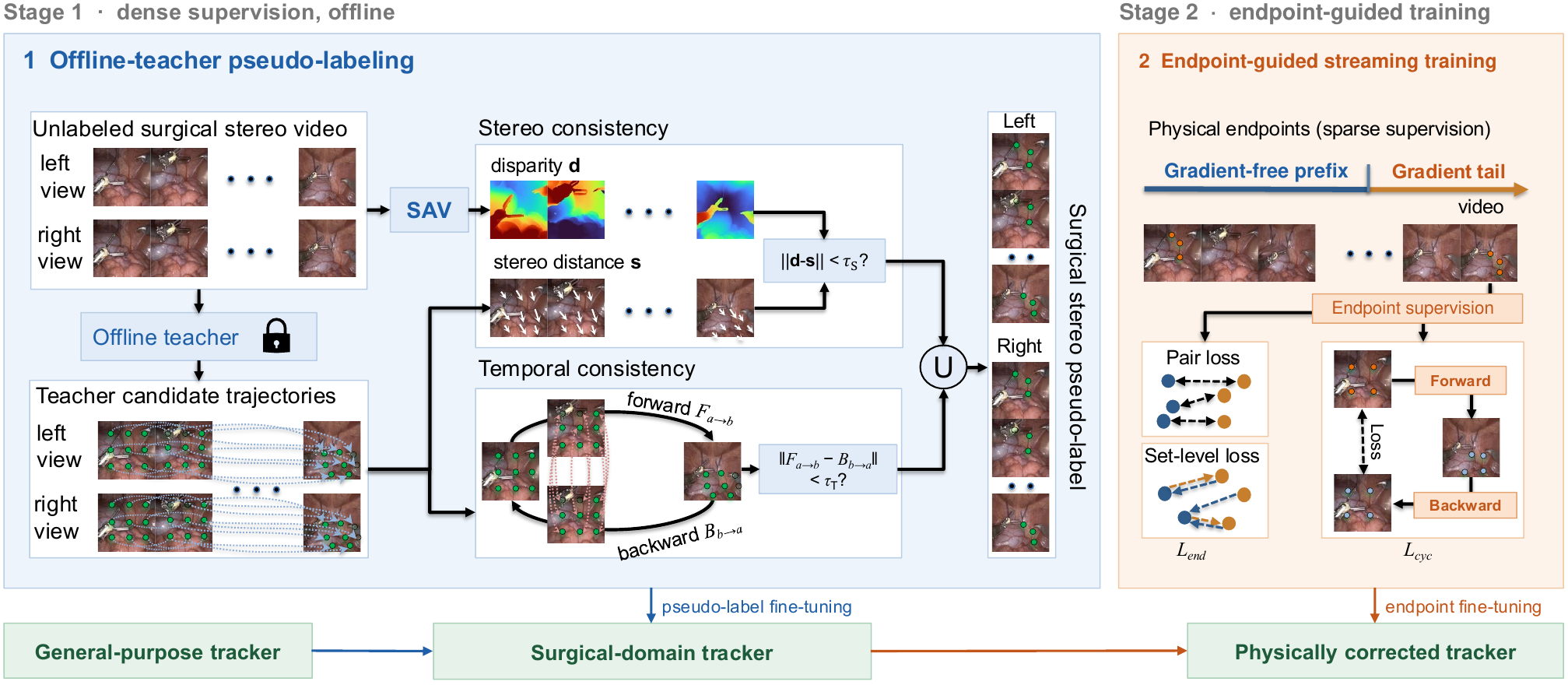}
\caption{\textbf{STEP framework.} (1) Temporal and stereo checks select training trajectories. (2) Endpoint-guided training combines physical endpoint and round-trip losses, with gradients limited to a final window.}
\label{fig:method}\label{fig:modules}
\end{figure}

The second stage, endpoint-guided training (S2), combines
a physical endpoint loss $\mathcal L_{\mathrm{end}}$ and a round-trip consistency
loss $\mathcal L_{\mathrm{cyc}}$, also called the cycle loss. Given a clip
$V=\{I_t\}_{t=1}^{T}$ and query points $\mathcal P_1$, we adapt $\theta_1$
using measured endpoints $\mathcal P_T$ and forward--backward tracks.
Both losses update the tracker before deployment (Figure~\ref{fig:method}).
VL-SurgPT supplies paired endpoint identities; STIR supplies unordered endpoint
sets whose cardinalities can differ.

\subsection{Streaming Training with a Bounded Gradient Window}
Training uses the tracker's causal forward pass. Let $h_t$ denote its recurrent
state, including point memory. Each new frame updates the state and predictions:
\begin{equation}
(h_t,\widehat{\mathcal P}_t,\widehat v_t)
=f_\theta(I_t,h_{t-1};\mathcal P_1).
\label{eq:stream}
\end{equation}
An endpoint label may be hundreds of frames from its query, whereas storing
activations for the entire clip is expensive. We therefore use a truncated backpropagation-through-time approximation~\citep{werbos1990bptt,williams1990online}. We split the clip at $s=T-B$,
stream frames $1{:}s$ without gradients, and use the predicted positions to
initialize a differentiable $B$-frame tail:
\begin{equation}
\overline{\mathcal P}_s=\operatorname{stopgrad}
 \bigl(\operatorname{Stream}_{\theta}(I_{1:s},\mathcal P_1)\bigr),
\qquad
\widehat{\mathcal P}_T=G_\theta(I_{s+1:T};\overline{\mathcal P}_s).
\label{eq:tail}
\end{equation}
Here $G_\theta$ shares the tracker weights. For Track-On-R, the tail transfers predicted positions and initializes fresh
point memory and query descriptors at its boundary. Deployment maintains
continuous streaming memory across frames.
This truncated training approximation exposes the tail to accumulated drift
while limiting stored temporal activations to $B$ frames.

\textbf{Learning from accumulated error.}
Initializing the tail from the student's prediction preserves the displacement
error accumulated over the prefix. Reinitializing it from a ground-truth point
would instead train recovery from an accurate boundary location. Our construction
exposes the final window to the former condition and computes endpoint gradients
through that window. The prefix therefore supplies realistic starting positions
for correction, while the gradient horizon remains independent of
the full clip length.

A $K\times K$ support grid supplies context alongside the annotated queries.
Only annotated queries contribute to endpoint losses and evaluation.

\subsection{Physical Endpoint Loss}
\textbf{Paired endpoints.}
When identities are available, the prediction for query $i$ is compared directly
with its measured endpoint $p_i^T$:
\begin{equation}
\mathcal L_{\mathrm{end}}^{\mathrm{pair}}
=\frac{1}{N}\sum_{i=1}^{N}\rho_\delta(\|\widehat p_i^T-p_i^T\|_2),
\label{eq:end-pair}
\end{equation}
where $\rho_\delta$ is the Huber robust penalty~\citep{huber1964}, with $\rho_\delta(d)=d^2/(2\delta)$ for $d\leq\delta$ and
$\rho_\delta(d)=d-\delta/2$ otherwise. We use $\delta=8$ pixels in the
$640\times512$ training-coordinate system. The linear tail limits the influence
of a widely displaced prediction while preserving a positional correction.
Only valid physical endpoints contribute to the average.

Physical observations can contain localization noise, particularly when marker
acquisition and white-light frames are not perfectly aligned. The Huber penalty
provides quadratic refinement near a measured point and a bounded derivative
for large residuals. This retains a correction signal for displaced predictions
without letting one endpoint dominate the update.

\textbf{Unordered endpoints.}
For observations without persistent start-to-end identities, we use a bidirectional robust
nearest-neighbor distance between predicted and observed endpoint sets, analogous to Chamfer-style point-set objectives~\citep{fan2017pointset}:
\begin{equation}
\mathcal L_{\mathrm{end}}^{\mathrm{set}}=
\frac{1}{|\widehat{\mathcal P}_T|}\sum_{\widehat p\in\widehat{\mathcal P}_T}
\rho_\delta\!\left(\min_{p\in\mathcal P_T}\|\widehat p-p\|_2\right)
+\frac{\beta}{|\mathcal P_T|}\sum_{p\in\mathcal P_T}
\rho_\delta\!\left(\min_{\widehat p\in\widehat{\mathcal P}_T}\|p-\widehat p\|_2\right).
\label{eq:end-set}
\end{equation}
The prediction-to-observation term aligns tracks with measured markers; the
reverse term, weighted by $\beta=0.5$, encourages coverage of the observed set.
This accommodates unequal set sizes. When endpoint identities are supplied,
Equation~\ref{eq:end-pair} preserves them: a permutation of two correctly located
markers can have zero set distance while still assigning each query incorrectly.

\subsection{Round-Trip Consistency Loss}
The endpoint loss anchors the final position. To constrain temporal consistency~\citep{wang2019timecycle,jabri2020crw},
we track auxiliary queries $c_j$ forward and then through the reversed clip.
Writing $\widehat c_j$ for the returned location gives
\begin{equation}
\mathcal L_{\mathrm{cyc}}=\frac{1}{M}\sum_{j=1}^{M}\rho_\delta(\|\widehat c_j-c_j\|_2),
\qquad
\mathcal L_{\mathrm{stream}}=\mathcal L_{\mathrm{end}}+
\lambda\mathcal L_{\mathrm{cyc}}.
\label{eq:cycle}
\end{equation}
The forward traversal runs without gradients. Queries whose terminal positions
leave the image are removed, and the return pass retains a gradient tail.
Physical endpoints anchor tissue correspondence, while the cycle loss
encourages return consistency along the predicted path.
The joint Track-On-R recipe accumulates five endpoint-labeled clips and one cycle clip per
optimizer step with unit per-clip loss weights, giving an effective expected
relative weight $\lambda=1/5$. Cycle clips contain 96 frames with a 12-frame
gradient tail; auxiliary queries use an $8\times8$ grid.

The cycle objective supplies training queries beyond the sparsely marked tissue
points. Its target is the known starting location, so no additional endpoint
annotation is required for these auxiliary tracks. Physical endpoints and cycles
are optimized together: measurements identify where tracked tissue should end,
while round trips test the consistency of the student's predicted correspondence.

\textbf{Inference.}
After both training stages, weights remain fixed. Given initial query coordinates,
the tracker processes RGB frames in a single forward stream with continuous
memory (Equation~\ref{eq:stream}). Only training requires teacher predictions, endpoint labels, stereo inputs, and reverse tracking.

\section{Experiments}
\label{sec:exp}
\subsection{Experimental Setup}
\textbf{Training.}
We adapt released Track-On, CoTracker3, and Track-On-R checkpoints
\citep{aydemir2025trackon,karaev2025cotracker3,aydemir2026trackonr}.
Track-On-R Stage~1 uses eight 48-frame clips per optimizer step and a 32-frame
gradient window. Its step-100 checkpoint initializes Stage~2. Both stages use
AdamW with a frozen image encoder, peak learning rate $3\times10^{-6}$, and
weight decay $10^{-5}$~\citep{kingma2015adam,loshchilov2019adamw}.
Track-On and CoTracker3 combine the endpoint objective with their native dense
trajectory losses; Track-On-R uses endpoint and round-trip losses.
Appendix~\ref{app:implementation} gives the optimization schedule and window sizes.

\textbf{Evaluation splits.}
We evaluate curation by training on STIR and transferring to all 150 VL-SurgPT
clips (375 physical endpoints). Endpoint-guided training uses a separate
603-clip manifest that overlaps with 121 of these clips. The remaining 29 clips
(73 endpoints) form a post-hoc clip-disjoint diagnostic, used for the main
Track-On-R training comparison. It excludes those clips from endpoint training
but does not establish procedure-level independence or a prospectively reserved
test set. Tables~\ref{tab:main} and~\ref{tab:occlusion} describe the full
150-clip pool; Table~\ref{tab:compact} uses the 29-clip diagnostic.
Appendix~\ref{app:evaluation} details the protocols. These comparisons jointly vary label source, density, and placement.

\textbf{Endpoint metrics.}
Following threshold-averaged TAP evaluation~\citep{doersch2022tapvid}, for $N$ evaluated endpoints with errors $e_i$, we report
\begin{equation}
\delta_{\mathrm{avg}}=\frac{100}{N|\mathcal A|}
\sum_{i=1}^{N}\sum_{a\in\mathcal A}\mathbf 1[e_i\leq a],
\qquad \mathrm{EPE}=\frac{1}{N}\sum_{i=1}^{N}e_i.
\label{eq:metrics}
\end{equation}
VL-SurgPT uses paired identities, coordinates scaled to $720\times480$, and
$\mathcal A=\{2,4,8,16,32\}$ pixels. Points are pooled across clips using the
same valid point indices for each method. STIR uses 60 clips and 496 predicted
endpoints, native $1280\times1024$ coordinates, and thresholds
$\{4,8,16,32,64\}$; its legacy $\delta_{60}$ uses prediction-to-nearest-marker
matching. This score measures set proximity, whereas VL-SurgPT measures identity.
Intermediate frames and terminal observations are evaluated separately.

\textbf{Statistical analysis.}
For the 29-clip diagnostic, we use a nonparametric paired bootstrap~\citep{efron1979}: we resample clips jointly across methods 10,000 times,
pool their endpoint errors, and report percentile 95\% intervals. This preserves
within-clip dependence. Intervals describe variation over clips for the saved
checkpoints; they do not incorporate training-seed variation.

\subsection{Physical Endpoint Tracking}
\label{sec:main}\label{sec:stir}\label{sec:support}
\begin{table}[t]\centering\small
\caption{\textbf{VL-SurgPT physical-endpoint comparison.}
STEP is compared with released models and manual-trajectory fine-tuning across multiple tracking backbones.}
\label{tab:main}
\setlength{\tabcolsep}{5pt}
\renewcommand{\arraystretch}{1.0}
\begin{tabular}{llrr}
\toprule
Backbone & Supervision
& $\delta_{\mathrm{avg}}\uparrow$
& EPE$\downarrow$ (px) \\
\midrule

RAFT~\citep{teed2020raft}
& none  & 33.81 & 82.16 \\

SEA-RAFT~\citep{wang2024searaft}
& none  & 33.71 & 80.17 \\

TAPIR~\citep{doersch2023tapir}
& none  & 64.80 & 15.64 \\

\midrule

\multirow{3}{*}{Track-On~\citep{aydemir2025trackon}}
& none            & 67.36 & 11.79 \\
& manual trajectories & 73.92 & 11.05 \\
\rowcolor{highlight}
& STEP (Ours)
& \textbf{75.89}\improve{+1.97}
& \textbf{10.70}\improve{$-$0.35} \\

\midrule

\multirow{3}{*}{CoTracker3~\citep{karaev2025cotracker3}}
& none            & 65.87 & 12.14 \\
& manual trajectories & 66.93 & 13.08 \\
\rowcolor{highlight}
& STEP (Ours)
& \textbf{67.52}\improve{+0.59}
& \textbf{11.33}\improve{$-$1.75} \\

\midrule

& none            & 72.27 & 15.27 \\
& manual trajectories & 73.17 & 14.90 \\
\rowcolor{highlight}
\multirow{-3}{*}{Track-On-R~\citep{aydemir2026trackonr}}
& STEP (Ours)
& \textbf{74.08}\improve{+0.91}
& \textbf{12.14}\improve{$-$2.76} \\

\bottomrule
\end{tabular}
\end{table}
\textbf{Endpoint correspondence.}
On the 29-clip diagnostic, STEP reduces Track-On-R endpoint error from 31.52 to
16.80 pixels relative to dense-label fine-tuning, a 46.7\% reduction.
Table~\ref{tab:main} complements this comparison with results across three
backbones on the full evaluation pool. Endpoint-guided training improves both
metrics over dense-label fine-tuning for each backbone: EPE decreases from
11.05 to 10.70 pixels for Track-On, 13.08 to 11.33 for CoTracker3, and 14.90 to
12.14 for Track-On-R. These full-pool results describe adaptation across
architectures under the overlap reported in Section~\ref{sec:exp}.

\begin{table}[t]\centering\small
\caption{\textbf{STIR resolution and support-grid controls.}
STEP is compared with Track-On-R under matched inference settings.}
\label{tab:stir-controls}
\setlength{\tabcolsep}{10pt}
\begin{tabular}{llrr}\toprule
Setting & Working res. & Grid & $\delta_{60}\uparrow$ \\\midrule
Track-On-R & $384\times512$ & none  & 75.89 \\
Track-On-R & $384\times512$ & $8\times8$   & 76.53 \\
Track-On-R & $384\times512$ & $16\times16$ & 77.54 \\
Track-On-R & $480\times608$ & none  & 77.86 \\
Track-On-R & $480\times608$ & $8\times8$   & 78.10 \\
Track-On-R & $480\times608$ & $16\times16$ & 78.35 \\
\textbf{STEP} & $480\times608$ & $16\times16$ & \textbf{79.23} \\\bottomrule
\end{tabular}\end{table}

\textbf{STIR and inference controls.}
Table~\ref{tab:stir-controls} separates endpoint training from changes in
inference resolution and spatial support. Increasing resolution and adding a
support grid improve the released Track-On-R baseline. At the matched
$480\times608$ resolution with a $16\times16$ grid, STEP reaches
79.23 $\delta_{60}$ versus 78.35 for the unadapted model. Its EPE is 12.44 pixels,
compared with 16.37 for the released reference configuration. The matched
accuracy comparison supports a benefit from endpoint training beyond inference
settings. Appendix~\ref{app:stir3d} evaluates the same predictions in 3D:
STEP reaches 74.71 average accuracy versus 51.48 for the static start-cloud
control. This measures recovery of tissue displacement relative to leaving
points at their initial 3D positions.

\subsection{Effect of Data Curation}
\label{sec:curation-results}\label{sec:ab-pseudo}
\begin{table}[t]
\centering\small
\begin{minipage}[b]{0.51\linewidth}
\caption{\textbf{Consistency selection.} STIR training; VL-SurgPT evaluation. EPE in pixels.}
\label{tab:pseudo}
\centering\setlength{\tabcolsep}{2pt}
\renewcommand{\arraystretch}{1.10}
\begin{tabular*}{\linewidth}{@{\extracolsep{\fill}}lrrr@{}}
\toprule
Selection & Keep (\%) & $\delta_{\mathrm{avg}}\uparrow$ & EPE$\downarrow$ \\
\midrule
Unfiltered & 100.0 & 72.37 & 15.99 \\
Temporal (T) & 57.2 & 72.16 & 16.40 \\
Stereo (S) & 52.8 & 72.48 & 15.96 \\
T\,$\cap$\,S & 47.5 & 71.04 & 20.42 \\
Count-matched T & 47.5 & 71.47 & 17.31 \\
\textbf{T\,$\cup$\,S} & 62.5 & \textbf{72.91} & \textbf{14.87} \\
\bottomrule
\end{tabular*}\par\vspace{0pt}
\end{minipage}\hfill
\begin{minipage}[b]{0.47\linewidth}
\caption{\textbf{Training-stage and loss controls.} 29-clip diagnostic; joint training also changes endpoint exposure and gradient-tail length. EPE is in pixels.}
\label{tab:compact}
\centering\setlength{\tabcolsep}{2pt}
\renewcommand{\arraystretch}{1.10}
\begin{tabular*}{\linewidth}{@{\extracolsep{\fill}}lrr@{}}
\toprule
Recipe & $\delta_{\mathrm{avg}}\uparrow$ & EPE$\downarrow$ \\
\midrule
S1 only & 68.77 & 19.91 \\
S2 endpoint only, no S1 & 70.96 & 20.08 \\
S1 + endpoint & 70.96 & 22.26 \\
S1 + cycle only & 70.68 & 23.38 \\
\textbf{S1 + endpoint + cycle} & \textbf{71.78} & \textbf{16.80} \\
\bottomrule
\end{tabular*}\par\vspace{0pt}
\end{minipage}
\end{table}

We compare six selection policies on the same STIR candidate pool, with identical
300-step optimization budgets and no VL-SurgPT adaptation (Table~\ref{tab:pseudo}).
The temporal--stereo union transfers at 72.91 accuracy and 14.87-pixel EPE,
compared with 72.37 and 15.99 for unfiltered candidates. It also exceeds the
released Track-On-R initialization's 72.27 accuracy. Relative to unfiltered
training, the union improves accuracy by 0.54 points and reduces EPE by 7.0\%.

\textbf{Selection criterion and retained count.}
The intersection retains 47.5\% of candidates in the source-pool probe and
transfers at 71.04 accuracy. Randomly reducing temporal candidates to the same
per-window count yields 71.47, with EPE 17.31 instead of 20.42 pixels.
At this matched count, requiring agreement with both tests does not improve
transfer. Complementary coverage is more effective here than requiring both tests. Union-versus-temporal comparisons additionally change
the number of retained candidates; they assess the complete selection policy.

Union selection also improves median error from 2.89 to 2.84 pixels and success
within 8 pixels from 83.47\% to 84.27\%. Both union and unfiltered training have
4.8\% of endpoints above 100 pixels, indicating improvements beyond the count
of extreme failures.

\subsection{Training Stages and Loss Design}
\label{sec:analysis}\label{sec:ab-stage}\label{sec:ab-loss}
\textbf{Curated initialization and physical supervision.}
Table~\ref{tab:compact} compares curated-data training (S1) and endpoint-only training (S2). On the diagnostic subset,
S1 gives 68.77 accuracy and 19.91-pixel EPE, while S2 without S1 gives 70.96
and 20.08. Following S1 with the endpoint loss also gives 70.96 accuracy, with EPE 22.26.
These controls show that curated initialization alone does not guarantee better
endpoint accuracy; the strongest result uses the joint training objective.

Cycle-only training after S1 reaches 70.68 accuracy and 23.38-pixel EPE.
Compared with S1 alone, its higher thresholded accuracy accompanies a higher
mean error, showing that these metrics capture different aspects of the error
distribution. The joint configuration improves both. This pattern supports using
measured terminal correspondence alongside return consistency to guide training.

\textbf{Loss combination.}
From the same S1 initialization, the joint recipe reaches 71.78 accuracy and 16.80-pixel EPE, versus 70.96 and 22.26 for endpoint-only training. The paired EPE-difference interval is $[-14.85,-0.11]$ pixels, whereas the accuracy interval $[-0.56,2.78]$ includes zero. The joint recipe also increases endpoint exposure and gradient-tail length
(Table~\ref{tab:recipes}), so this comparison measures their combined effect.

\begin{figure}[t]
\centering
\includegraphics[width=0.95\linewidth]{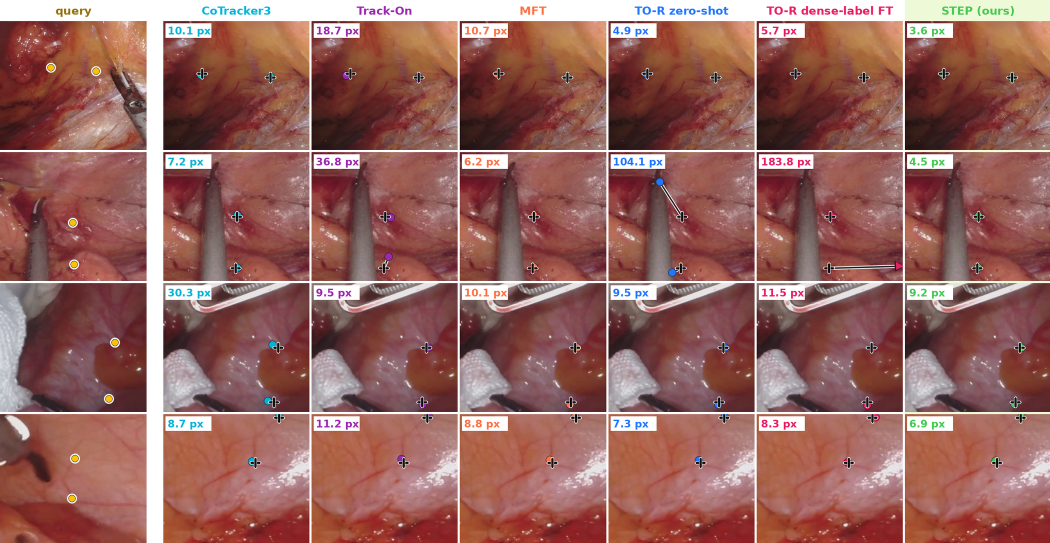}
\caption{\textbf{Qualitative endpoint tracking.} Predictions and physical references (crosses) on four VL-SurgPT sequences; errors are in pixels. The last three columns use the Track-On-R backbone.}
\label{fig:qual-methods}
\end{figure}

\subsection{Occlusion and Error Distribution}
\label{sec:robustness}\label{sec:intermediate}
\begin{table}
\caption{\textbf{Occlusion recovery and visibility.} Full-pool endpoints grouped by intermediate visibility. EPE is in pixels; other metrics are percentages.}
\label{tab:occlusion}
\vspace{-1mm}
\centering
\scriptsize
\setlength{\tabcolsep}{2.5pt}
\resizebox{\linewidth}{!}{%
\begin{tabular}{lcccccccc}
\toprule
& \multicolumn{3}{c}{Occluded mid-sequence ($n{=}145$)}
& \multicolumn{3}{c}{Never occluded ($n{=}230$)}
& \multicolumn{2}{c}{Visibility} \\
\cmidrule(lr){2-4}\cmidrule(lr){5-7}\cmidrule(lr){8-9}
Model & $\delta\uparrow$ & EPE$\downarrow$ & $>$100\,px$\downarrow$
& $\delta\uparrow$ & EPE$\downarrow$ & $>$100\,px$\downarrow$
& Occ.\ recall$\uparrow$ & False occ.$\downarrow$ \\
\midrule
Zero-shot       & 64.14 & 28.54 & 8.97 & 77.39 & 6.91 & 1.30 & 47.9 & 2.0 \\
Dense-label FT  & 64.28\improve{+0.14} & 25.92 & 9.66 & \textbf{78.78}\improve{+1.39} & 7.95 & 1.30 & 49.3 & \textbf{1.7} \\
Stage~1 only    & 64.55\improve{+0.41} & 28.51 & 8.28 & 77.74\improve{+0.35} & \textbf{6.53} & \textbf{0.87} & 49.9 & 1.9 \\
Stage~2 only    & 66.76\improve{+2.62} & 24.15 & \textbf{7.59} & 78.17\improve{+0.78} & 7.29 & 1.30 & 50.1 & 1.9 \\
\rowcolor{highlight}
STEP (ours)            & \textbf{66.76}\improve{+2.62} & \textbf{20.83} & \textbf{7.59} & 78.70\improve{+1.31} & 6.67 & 1.30 & \textbf{51.6} & 2.1 \\
\bottomrule
\end{tabular}}
\vspace{-1mm}
\end{table}

Table~\ref{tab:occlusion} groups the full-pool endpoints by intermediate visibility.
For points occluded mid-sequence, STEP improves accuracy from 64.28 to 66.76 and
reduces EPE from 25.92 to 20.83 pixels relative to dense-label fine-tuning.
Errors above 100 pixels decrease from 9.66\% to 7.59\%. For never-occluded points,
accuracy is similar (78.70 versus 78.78), with EPE decreasing from 7.95 to
6.67 pixels. The larger gain after occlusion supports endpoint supervision for correspondence recovery.

Visibility prediction provides a complementary view: STEP reaches 51.6\%
occlusion recall with 2.1\% false occlusion. Recovery at a visible endpoint and
recognition of an intermediate occlusion measure different behaviors.
Appendices~\ref{app:intermediate} and~\ref{app:scenario} report intermediate
tracking, acquisition conditions, and visibility on the 29-clip diagnostic.

\textbf{Intermediate correspondence.}
The annotated path provides another view of supervision. Across 2,293
intermediate point-frames, the three-seed S1+endpoint control improves accuracy
by 1.40 points over the released model and by 0.92 over S1, while dense-label
fine-tuning scores 1.62 points higher than this endpoint-only control
(Appendix~\ref{app:intermediate}). Physical endpoints can therefore improve
intermediate tracking as well as terminal correspondence, although direct
trajectory labels remain useful for supervising the intervening path.

\textbf{Efficiency.}
At $480\times608$ with a $16\times16$ support grid, Track-On-R STEP takes
63.3--66.0\,ms per frame and 1.21--1.37\,GB peak inference memory for 8--64 queries
on an A100-80GB. This corresponds to approximately 15--16 frames/s under the same
resolution and support-grid setting used for the STIR comparison. For a fixed
query count, streaming maintains constant state size.
Appendix~\ref{app:cost} reports peak memory usage during training.

\textbf{Qualitative behavior.}
Figure~\ref{fig:qual-methods} illustrates localized errors and larger displacements
to different surfaces. STEP stays closer to the physical references in these
examples. Avoiding large errors can improve EPE more than thresholded accuracy.
On four selected clips, STEP also produces fewer orientation flips in a tracked
grid than dense-label fine-tuning (Figure~\ref{fig:mesh}). These flips expose
local reversals in the predicted motion field that are difficult to see from
isolated endpoint markers. The grid and endpoint views assess spatial
organization and terminal accuracy together, providing complementary descriptions
of tracking behavior.

\FloatBarrier
\section{Conclusion}
STEP learns surgical correspondence from two complementary resources: dense
trajectories curated from video and sparse physical endpoints. Temporal and
stereo selection improve cross-dataset transfer, while endpoint-guided streaming
training exposes the tracker to accumulated error with bounded activation memory.
Combining physical endpoint and round-trip objectives yields lower endpoint error
on the VL-SurgPT diagnostic. These findings support using existing surgical
video and measured correspondence together, with dense targets providing broad
training coverage and physical observations anchoring long-range predictions.
The resulting tracker runs on a single RGB stream with fixed weights.

\clearpage
\section*{AI Use Statement}
Generative AI assisted with code implementation, manuscript editing, figure layout,
and consistency checks. The authors developed the research ideas, ran the experiments,
verified the reported results, and remain responsible for all content.

\section*{Ethics Statement}
This work uses existing surgical datasets under their access terms, with no new patient acquisition. The evaluated trackers are
research systems; clinical deployment would require further validation.

\section*{Reproducibility Statement}
Sections~\ref{sec:data} and~\ref{sec:method} specify the curation pipeline and
training objectives. Appendix~\ref{app:implementation} reports implementation
and efficiency settings; Appendices~\ref{app:evaluation} and~\ref{app:results}
detail evaluation protocols and additional results. Experiments initialize from released
tracker checkpoints and use existing surgical datasets.

\bibliographystyle{iclr2027_conference}
\bibliography{references}

@inproceedings{doersch2022tapvid,
  title     = {{TAP}-Vid: A Benchmark for Tracking Any Point in a Video},
  author    = {Doersch, Carl and Gupta, Ankush and Markeeva, Larisa and Recasens, Adria and
               Smaira, Lucas and Aytar, Yusuf and Carreira, Joao and Zisserman, Andrew and Yang, Yi},
  booktitle = {Advances in Neural Information Processing Systems (NeurIPS)},
  volume    = {35},
  pages     = {13610--13626},
  year      = {2022},
}

@inproceedings{doersch2023tapir,
  title     = {{TAPIR}: Tracking Any Point with Per-Frame Initialization and Temporal Refinement},
  author    = {Doersch, Carl and Yang, Yi and Vecerik, Mel and Gokay, Dilara and Gupta, Ankush and
               Aytar, Yusuf and Carreira, Joao and Zisserman, Andrew},
  booktitle = {Proceedings of the IEEE/CVF International Conference on Computer Vision (ICCV)},
  pages     = {10061--10072},
  year      = {2023},
}

@inproceedings{doersch2024bootstap,
  title     = {{BootsTAP}: Bootstrapped Training for Tracking-Any-Point},
  author    = {Doersch, Carl and Luc, Pauline and Yang, Yi and Gokay, Dilara and Koppula, Skanda and
               Gupta, Ankush and Heyward, Joseph and Rocco, Ignacio and Goroshin, Ross and
               Carreira, Jo{\~a}o and Zisserman, Andrew},
  booktitle = {Proceedings of the Asian Conference on Computer Vision (ACCV)},
  year      = {2024},
}

@inproceedings{karaev2025cotracker3,
  title     = {{CoTracker3}: Simpler and Better Point Tracking by Pseudo-Labelling Real Videos},
  author    = {Karaev, Nikita and Makarov, Iurii and Wang, Jianyuan and Neverova, Natalia and
               Vedaldi, Andrea and Rupprecht, Christian},
  booktitle = {Proceedings of the IEEE/CVF International Conference on Computer Vision (ICCV)},
  year      = {2025},
}

@inproceedings{aydemir2025trackon,
  title     = {{Track-On}: Transformer-based Online Point Tracking with Memory},
  author    = {Aydemir, G{\"o}rkay and Cai, Xiongyi and Xie, Weidi and G{\"u}ney, Fatma},
  booktitle = {International Conference on Learning Representations (ICLR)},
  year      = {2025},
}

@inproceedings{aydemir2026trackonr,
  title     = {Real-World Point Tracking with Verifier-Guided Pseudo-Labeling},
  author    = {Aydemir, G{\"o}rkay and G{\"u}ney, Fatma and Xie, Weidi},
  booktitle = {Proceedings of the IEEE/CVF Conference on Computer Vision and Pattern Recognition (CVPR)},
  year      = {2026},
}

@inproceedings{neoral2024mft,
  title     = {{MFT}: Long-Term Tracking of Every Pixel},
  author    = {Neoral, Michal and {\v{S}}er{\'y}ch, Jon{\'a}{\v{s}} and Matas, Ji{\v{r}}{\'i}},
  booktitle = {Proceedings of the IEEE/CVF Winter Conference on Applications of Computer Vision (WACV)},
  pages     = {6837--6847},
  year      = {2024},
}

@inproceedings{teed2020raft,
  title     = {{RAFT}: Recurrent All-Pairs Field Transforms for Optical Flow},
  author    = {Teed, Zachary and Deng, Jia},
  booktitle = {Proceedings of the European Conference on Computer Vision (ECCV)},
  year      = {2020},
}

@inproceedings{wang2024searaft,
  title     = {{SEA-RAFT}: Simple, Efficient, Accurate {RAFT} for Optical Flow},
  author    = {Wang, Yihan and Lipson, Lahav and Deng, Jia},
  booktitle = {Proceedings of the European Conference on Computer Vision (ECCV)},
  year      = {2024},
}

@article{schmidt2024stir,
  title   = {Surgical Tattoos in Infrared: A Dataset for Quantifying Tissue Tracking and Mapping},
  author  = {Schmidt, Adam and Mohareri, Omid and DiMaio, Simon and Salcudean, Septimiu E.},
  journal = {IEEE Transactions on Medical Imaging (TMI)},
  year    = {2024},
}

@inproceedings{zhou2026vlsurgpt,
  title     = {Bridging Vision and Language for Robust Context-Aware Surgical Point Tracking:
               The {VL-SurgPT} Dataset and Benchmark},
  author    = {Zhou, Rulin and He, Wenlong and Wang, An and Zhang, Jianhang and Zeng, Xuanhui and
               Zhang, Xi and Zhu, Chaowei and Hu, Haijun and Ren, Hongliang},
  booktitle = {Proceedings of the AAAI Conference on Artificial Intelligence (AAAI)},
  year      = {2026},
}

@inproceedings{zhan2025surgmotion,
  title     = {Tracking Everything in Robotic-Assisted Surgery},
  author    = {Zhan, Bohan and Zhao, Wang and Fang, Yi and Du, Bo and Vasconcelos, Francisco and
               Stoyanov, Danail and Elson, Daniel S. and Huang, Baoru},
  booktitle = {IEEE International Conference on Robotics and Automation (ICRA)},
  year      = {2025},
}

@article{hayoz2023stereomis,
  title   = {Learning How To Robustly Estimate Camera Pose in Endoscopic Videos},
  author  = {Hayoz, Michel and Hahne, Christopher and Gallardo, Mathias and Candinas, Daniel and
             Kurmann, Thomas and Allan, Maximilian and Sznitman, Raphael},
  journal = {International Journal of Computer Assisted Radiology and Surgery (IJCARS)},
  year    = {2023},
  doi     = {10.1007/s11548-023-02919-w},
}

@misc{hamlyn,
  title        = {Hamlyn Centre Laparoscopic / Endoscopic Video Datasets},
  author       = {{Hamlyn Centre, Imperial College London}},
  howpublished = {\url{https://hamlyn.doc.ic.ac.uk/vision/}},
}

@inproceedings{godard2017stereo,
  title     = {Unsupervised Monocular Depth Estimation with Left-Right Consistency},
  author    = {Godard, Cl{\'e}ment and Mac Aodha, Oisin and Brostow, Gabriel J.},
  booktitle = {Proceedings of the IEEE Conference on Computer Vision and Pattern Recognition (CVPR)},
  year      = {2017},
  pages     = {270--279},
}

@inproceedings{schmidt2023sendd,
  title     = {{SENDD}: Sparse Efficient Neural Depth and Deformation for Tissue Tracking},
  author    = {Schmidt, Adam and Mohareri, Omid and DiMaio, Simon and Salcudean, Septimiu E.},
  booktitle = {Medical Image Computing and Computer Assisted Intervention (MICCAI)},
  year      = {2023},
  pages     = {238--248},
  doi       = {10.1007/978-3-031-43996-4_23},
}

@inproceedings{guo2024adatracker,
  title     = {Ada-Tracker: Soft Tissue Tracking via Inter-Frame and Adaptive-Template Matching},
  author    = {Guo, Jiaxin and Wang, Jiangliu and Li, Zhaoshuo and Jia, Tongyu and Dou, Qi and Liu, Yun-Hui},
  booktitle = {IEEE International Conference on Robotics and Automation (ICRA)},
  year      = {2024},
  pages     = {15463--15470},
  doi       = {10.1109/ICRA57147.2024.10611030},
}

@article{schmidt2024review,
  title   = {Tracking and Mapping in Medical Computer Vision: A Review},
  author  = {Schmidt, Adam and Mohareri, Omid and DiMaio, Simon and Yip, Michael C. and Salcudean, Septimiu E.},
  journal = {Medical Image Analysis},
  volume  = {94},
  pages   = {103131},
  year    = {2024},
  doi     = {10.1016/j.media.2024.103131},
}

@inproceedings{jing2025stereoanyvideo,
  title     = {Stereo Any Video: Temporally Consistent Stereo Matching},
  author    = {Jing, Junpeng and Luo, Weixun and Mao, Ye and Mikolajczyk, Krystian},
  booktitle = {Proceedings of the IEEE/CVF International Conference on Computer Vision (ICCV)},
  year      = {2025},
}

@inproceedings{harley2025alltracker,
  title={Alltracker: Efficient dense point tracking at high resolution},
  author={Harley, Adam W. and You, Yang and Sun, Xinglong and Zheng, Yang and Raghuraman, Nikhil and Gu, Yunqi and Liang, Sheldon and Chu, Wen-Hsuan and Dave, Achal and Tokmakov, Pavel and You, Suya and Ambrus, Rares and Fragkiadaki, Katerina and Guibas, Leonidas J.},
  booktitle={Proceedings of the IEEE/CVF International Conference on Computer Vision (ICCV)},
  year={2025}
}

@inproceedings{meister2018unflow,
  title={Unflow: Unsupervised learning of optical flow with a bidirectional census loss},
  author={Meister, Simon and Hur, Junhwa and Roth, Stefan},
  booktitle={Proceedings of the AAAI Conference on Artificial Intelligence},
  volume={32},
  year={2018}
}

@inproceedings{wang2019timecycle,
  title={Learning correspondence from the cycle-consistency of time},
  author={Wang, Xiaolong and Jabri, Allan and Efros, Alexei A.},
  booktitle={Proceedings of the IEEE/CVF Conference on Computer Vision and Pattern Recognition (CVPR)},
  pages={2566--2576},
  year={2019}
}

@inproceedings{jabri2020crw,
  title={Space-time correspondence as a contrastive random walk},
  author={Jabri, Allan and Owens, Andrew and Efros, Alexei A.},
  booktitle={Advances in Neural Information Processing Systems (NeurIPS)},
  volume={33},
  year={2020}
}

@book{hartley2004mvg,
  title={Multiple View Geometry in Computer Vision},
  author={Hartley, Richard and Zisserman, Andrew},
  publisher={Cambridge University Press},
  edition={2},
  year={2004}
}

@article{werbos1990bptt,
  title={Backpropagation through time: What it does and how to do it},
  author={Werbos, Paul J.},
  journal={Proceedings of the IEEE},
  volume={78},
  number={10},
  pages={1550--1560},
  year={1990}
}

@article{williams1990online,
  title={An efficient gradient-based algorithm for on-line training of recurrent network trajectories},
  author={Williams, Ronald J. and Peng, Jing},
  journal={Neural Computation},
  volume={2},
  number={4},
  pages={490--501},
  year={1990}
}

@article{huber1964,
  title={Robust estimation of a location parameter},
  author={Huber, Peter J.},
  journal={The Annals of Mathematical Statistics},
  volume={35},
  number={1},
  pages={73--101},
  year={1964}
}

@inproceedings{fan2017pointset,
  title={A point set generation network for 3d object reconstruction from a single image},
  author={Fan, Haoqiang and Su, Hao and Guibas, Leonidas J.},
  booktitle={Proceedings of the IEEE Conference on Computer Vision and Pattern Recognition (CVPR)},
  pages={605--613},
  year={2017}
}

@inproceedings{kingma2015adam,
  title={Adam: A method for stochastic optimization},
  author={Kingma, Diederik P. and Ba, Jimmy},
  booktitle={International Conference on Learning Representations (ICLR)},
  year={2015}
}

@inproceedings{loshchilov2019adamw,
  title={Decoupled weight decay regularization},
  author={Loshchilov, Ilya and Hutter, Frank},
  booktitle={International Conference on Learning Representations (ICLR)},
  year={2019}
}

@inproceedings{smith2017clr,
  title={Cyclical learning rates for training neural networks},
  author={Smith, Leslie N.},
  booktitle={IEEE Winter Conference on Applications of Computer Vision (WACV)},
  pages={464--472},
  year={2017}
}

@article{smith2018superconvergence,
  title={Super-convergence: Very fast training of neural networks using large learning rates},
  author={Smith, Leslie N. and Topin, Nicholay},
  journal={arXiv preprint arXiv:1708.07120},
  year={2018}
}

@article{efron1979,
  title={Bootstrap methods: Another look at the jackknife},
  author={Efron, Bradley},
  journal={The Annals of Statistics},
  volume={7},
  number={1},
  pages={1--26},
  year={1979}
}
\clearpage
\appendix
\section{Data and Curation}
\label{app:data}

\begin{table}[htbp]\centering\small
\caption{\textbf{Surgical sources and generated trajectory inventory.} STIR counts both views of
566 stereo clips. Pseudo-labels are model-generated; physical labels are existing endpoint
observations.}
\label{tab:data}
\setlength{\tabcolsep}{10pt}
\begin{tabular}{lrr}\toprule
 & VL-SurgPT & STIR \\\midrule
Video streams                    & 754 & 1{,}132 \\
Frames                           & 181\,k & 640\,k \\
Acquisition                      & Monocular & Stereo \\
Human-annotated trajectories     & 1{,}862 & 0 \\
Generated pseudo-trajectories    & 493{,}893 & 1{,}685{,}211 \\
Pseudo-labeled point-frames      & 103\,M & 300\,M \\
Selection rule                   & Temporal & Temporal $\cup$ stereo \\
Physical supervision             & Paired endpoints & Endpoint sets \\\bottomrule
\end{tabular}\end{table}

\textbf{VL-SurgPT.} VL-SurgPT provides monocular surgical video, human-annotated
trajectories, and paired physical start/end observations~\citep{zhou2026vlsurgpt}.
Table~\ref{tab:data} summarizes the generated inventory. Temporal selection is
used because the source is monocular. Excluding the 150 evaluation clips before
Stage~1 window sampling leaves 604 source clips; 596 yield usable training windows.

\textbf{STIR.} STIR contains 566 stereo clips, represented as 1,132 camera-view
streams~\citep{schmidt2024stir}. Temporal reversibility and stereo geometry allow
four consistency policies: temporal, stereo, intersection, and union. The policy
comparison uses a common pool of 1,726 clip-view windows and 300 optimizer steps
per policy. Multiple windows can originate from the same source clip.

\textbf{Additional sources and preprocessing.} The pipeline also generates
trajectories for StereoMIS, Hamlyn, and SurgMotion
\citep{hayoz2023stereomis,hamlyn,zhan2025surgmotion}. StereoMIS trajectories are
aligned to the decoded packet timeline so that coordinates correspond to displayed
frames. Hamlyn uses temporal selection with a first-frame grid check to remove
static border candidates. These steps prevent timing offsets and invalid image
regions from entering the training targets. The reported quantitative studies
use STIR and VL-SurgPT.

\textbf{Trajectory generation.}
STIR and VL-SurgPT candidates are generated by frozen
AllTracker~\citep{harley2025alltracker} over windows of at most 750 frames. The teacher
occlusion score is $1-\mathrm{visibility}\cdot\mathrm{confidence}$; scores at most 0.4 are treated
as visible. Temporal selection uses a 3-pixel round-trip threshold. On stereo sources, Stereo Any
Video~\citep{jing2025stereoanyvideo} supplies disparity; track pairing uses a 3-pixel tolerance, with an 8-pixel disparity tolerance on visible frames.

\FloatBarrier
\section{Training and Efficiency}
\label{app:implementation}\label{app:cost}

\textbf{Stage 1 optimization.} The Track-On-R S1 run uses eight 48-frame clips per optimizer step, a
32-frame differentiated window, and 400 optimizer steps. The checkpoint at step 100 initializes
Stage~2. The trainable part contains 24.95M parameters; the image encoder is frozen. Optimization
uses AdamW~\citep{kingma2015adam,loshchilov2019adamw} with weight decay $10^{-5}$ and a one-cycle
schedule~\citep{smith2017clr,smith2018superconvergence}; the peak learning rate is
$3\times10^{-6}$ with 10\% warm-up. All STIR selection policies share the released initialization and 300-step budget.

\textbf{Stage 2 optimization.} The endpoint-only Track-On-R control accumulates two endpoint clips
per optimizer step, processes streams up to 192 frames, and differentiates the last 16 frames. The
joint endpoint+cycle recipe accumulates five endpoint clips and one cycle clip per optimizer step.
Endpoint clips can span up to 256 frames and use a 24-frame gradient tail. The cycle clip contains
96 frames, uses a 12-frame return-gradient tail, and seeds an $8\times8$ auxiliary grid. Per-clip
losses have unit weights, so the expected cycle-to-endpoint contribution induced by the sampling
schedule is $1/5$. Both recipes use 300 optimizer steps.

\begin{table}[htbp]\centering\small
\caption{\textbf{Track-On-R endpoint-guided training recipes.} Clips accumulate per optimizer step; all S2 runs use 300 steps. The cycle loss uses a separate 96-frame clip and 12-frame gradient tail.}
\label{tab:recipes}
\setlength{\tabcolsep}{7pt}
\begin{tabular}{lrrrr}\toprule
Losses & Endpoint clips & Cycle clips & Max.\ span & Grad.\ tail \\\midrule
Endpoint only & 2 & 0 & $\leq192$ & 16 \\
Endpoint + cycle & 5 & 1 & $\leq256$ & 24 \\\bottomrule
\end{tabular}\end{table}

\textbf{Training coordinates and losses.}
The endpoint Huber transition~\citep{huber1964} is $\delta=8$ pixels in the
$640\times512$ training coordinate system. The bidirectional nearest-neighbor
loss for unordered STIR endpoints uses $\beta=0.5$.

\textbf{Runtime and memory.}
Runtime is measured on an A100-80GB with raw $1280\times1024$ frames resized to the indicated
working resolution. For fixed query count and memory size, streaming inference maintains constant
per-frame state size and $\mathcal O(T)$ total computation. At $480\times608$ with a $16\times16$
support grid, the joint Track-On-R checkpoint takes 63.3--66.0\,ms per frame and 1.21--1.37\,GB peak
memory for 8--64 queries. Endpoint-only training uses 30.5\,GB peak memory with a 16-frame gradient tail.
This training measurement applies to the endpoint-only configuration in
Table~\ref{tab:recipes}.

\FloatBarrier
\section{Evaluation Protocols}
\label{app:evaluation}\label{app:splits}\label{app:metrics}\label{app:statistics}

\textbf{Evaluation populations.}
The VL-SurgPT evaluation pool contains 150 clips and 375 physical endpoints.
The endpoint-training manifest contains 603 clips, including 121 from this pool;
the remaining 29 clips and 73 endpoints form the post-hoc clip-disjoint diagnostic.
Full-pool analyses describe the existing checkpoints across backbones, visibility
groups, and acquisition conditions. The 29-clip subset evaluates clips absent
from endpoint training, although procedure independence and prior model-selection
exposure are not established. STIR-to-VL-SurgPT curation transfer uses the full
pool because those models are trained solely on STIR.

\textbf{Endpoint scoring.}
VL-SurgPT predictions and labels are scaled together to $720\times480$;
STIR's legacy image-plane endpoint score uses native $1280\times1024$
coordinates. Section~\ref{sec:exp} defines scoring and matching.

\textbf{STIR 3D protocol.} We follow the official STIR~2024 3D evaluation~\citep{schmidt2024stir}: each
predicted 3D endpoint is matched by nearest neighbour to the ground-truth end cloud, and accuracy
is the fraction of matched distances below a threshold, averaged over
$\{2,4,8,16,32\}$\,mm. Scoring uses 59 clips and 495 endpoints. The \emph{static} row is the
protocol's built-in control: the unchanged ground-truth \emph{start} cloud is the prediction, measuring accuracy without tracking.

\textbf{Statistical analysis.}
For the 29-clip diagnostic, we draw 10,000 bootstrap samples with replacement,
using the same clip identifiers for every method. Endpoint errors are pooled;
the 2.5 and 97.5 percentiles define the interval~\citep{efron1979}. Resampling
clips preserves within-clip dependence. These intervals describe clip variation
for fixed checkpoints, rather than training-seed variation.

\FloatBarrier
\section{Additional Results}
\label{app:results}

\subsection{Intermediate Tracking}
\label{app:intermediate}
We evaluate intermediate correspondence on 2,293 annotated point-frames from the
150-clip pool. These samples exclude both the initial query frame and the final
physical endpoint, measuring the path between them.

For the three-seed S1+endpoint control without the cycle loss, a paired clip-level bootstrap with
10,000 resamples gives a mean $\delta_{\mathrm{avg}}$ difference of $+1.40$ points relative to the
released initialization (95\% interval $[+0.37,+2.59]$), $+0.92$ relative to S1
($[+0.02,+1.81]$), and $-1.62$ relative to dense human-trajectory fine-tuning
($[-3.31,-0.02]$). Thus, endpoint-only training improves intermediate correspondence over the
released and S1 models, while dense trajectory supervision remains stronger
on the annotated path.

For normalized time, let
\begin{equation}
\tau=\frac{k-k_{\mathrm{first}}}{k_{\mathrm{last}}-k_{\mathrm{first}}},
\label{eq:normtime}
\end{equation}
where $k_{\mathrm{first}}$ is the first annotated frame after the query and $k_{\mathrm{last}}$ is
the endpoint. Four bins over $\tau>0$ contain 378, 483, 513, and 581 point-frames, that is 1,955 of the 2,293
intermediate samples; the remaining 338 lie at $\tau=0$ and fall outside the binned range. The 375 physical endpoints are scored separately from these intermediate samples.

\subsection{Acquisition Conditions}
\label{app:scenario}
We group full-pool endpoints by acquisition condition for the joint Track-On-R model. Physical endpoint $\delta_{\mathrm{avg}}$/EPE values are 74.93/14.59 for tissue
deformation ($n=71$), 75.07/13.66 for instrument occlusion ($n=69$), 73.49/15.31 for camera jitter
($n=86$), 66.88/8.27 for surface reflection ($n=64$), and 78.59/8.57 for cauterization smoke
($n=85$). Across all 375 endpoints, the joint checkpoint gives 74.08/12.14 and 3.73\% of errors
above 100 pixels. 

\textbf{Visibility on the clip-disjoint diagnostic.}
On the 29-clip subset, visibility is scored on 59 occluded and 569 visible
annotated point-frames. Occlusion recall
conditions on occluded observations; false-occlusion rate conditions on visible ones. STEP gives
50.85\% recall and 2.99\% false occlusion, while S1 gives 62.71\% and 3.51\%. Endpoint recovery and visibility recall measure different behavior.

\subsection{STIR 3D Accuracy}
\label{app:stir3d}
STIR's calibrated stereo observations provide 3D ground truth in addition to
image-plane endpoints. We evaluate the predictions from the main STIR experiment
with the 3D protocol in Appendix~\ref{app:metrics}, comparing STEP with the
static start-cloud control (Table~\ref{tab:stir3d}).

\begin{table}[htbp]\centering\small
\caption{\textbf{STIR 3D endpoint accuracy.} Official STIR 2024 3D protocol, 59 clips and 495
endpoints, nearest-neighbour matched to the ground-truth end cloud. Higher is better.}
\label{tab:stir3d}
\setlength{\tabcolsep}{8pt}
\begin{tabular}{lrrrrrr}\toprule
Prediction & 2\,mm & 4\,mm & 8\,mm & 16\,mm & 32\,mm & Avg$\uparrow$ \\\midrule
Static control (GT start cloud) & 15.85 & 29.55 & 49.68 & 71.09 & 91.22 & 51.48 \\
\rowcolor{highlight}\textbf{STEP} & \textbf{37.37} & \textbf{62.83} & \textbf{81.21} & \textbf{93.13} & \textbf{98.99} & \textbf{74.71} \\\bottomrule
\end{tabular}\end{table}

STEP reaches 74.71 average accuracy, compared with 51.48 for the static control.
The 23.23-point difference measures the benefit over leaving points at their
initial 3D locations. At 2\,mm, accuracy increases from 15.85 to 37.37; at 32\,mm,
both methods accept most endpoints (91.22 and 98.99). The tighter thresholds
therefore distinguish correspondence more strongly in this comparison.

\FloatBarrier
\section{Qualitative Analysis}
\label{app:qual}

\begin{figure}[htbp]
\centering
\includegraphics[width=\linewidth]{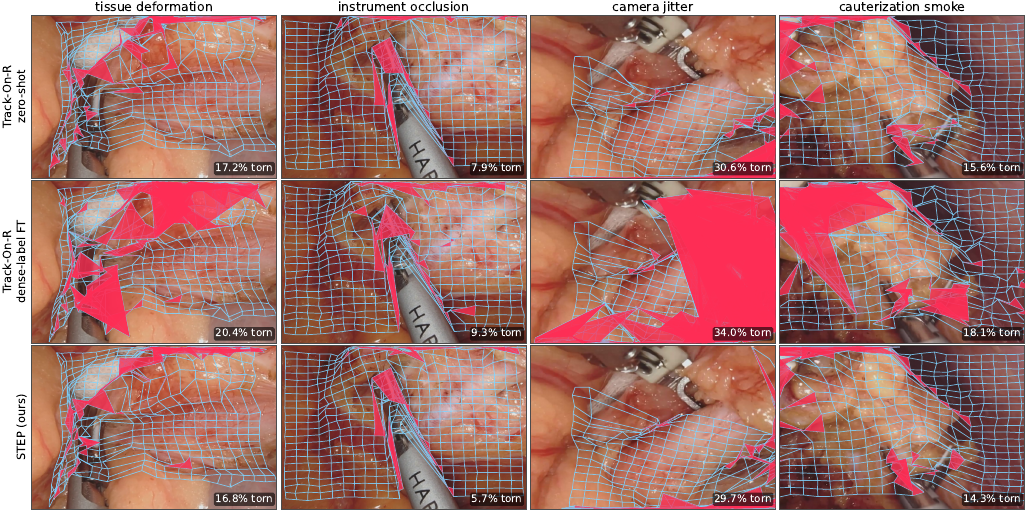}
\caption{\textbf{Motion-field coherence without ground truth.} Track-On-R variants on four VL-SurgPT clips. Red cells mark orientation flips and panels report their rates; compare within columns. Coherent tracks can still drift.}
\label{fig:mesh}
\end{figure}

\textbf{Spatial coherence.}
Figure~\ref{fig:mesh} tracks a regular grid from the first to the final frame of
four selected clips from the 29-clip diagnostic, one per represented VL-SurgPT
scenario. All three models use the Track-On-R backbone with different supervision. A grid cell is marked as a tear when its orientation flips
relative to the first frame. Dense-label fine-tuning has the highest tear rate
in each displayed scenario and STEP the lowest. Rates depend on grid density
and clip length, so comparisons are made within each column. The clips were
selected for visual interpretability.

Smooth mesh deformation is a spatial-coherence diagnostic. A field can drift
coherently to incorrect positions, while genuine motion discontinuities at
instrument boundaries can produce tears. Across the measured clips, tear rate
and endpoint error are uncorrelated (Spearman $\rho=-0.09$ over 21 clip-method
pairs). Spatial coherence and endpoint accuracy therefore capture complementary properties.

\begin{figure}[htbp]
\centering
\includegraphics[width=0.8\linewidth]{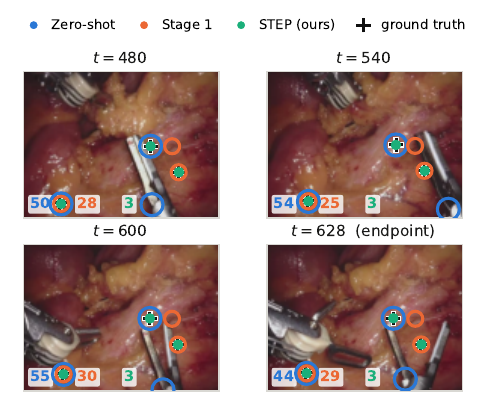}
\caption{\textbf{Additional correspondence examples.} Predictions at selected frames illustrate tracking through the sequence; numbers give errors in pixels.}
\label{fig:qual-extra}
\end{figure}

\textbf{Endpoint correspondence.}
Figure~\ref{fig:qual-extra} complements the endpoint comparison with predictions
at selected frames. The examples illustrate how correspondence evolves during
tissue motion and tool interaction; the quantitative endpoint and visibility
analyses provide the corresponding measured evaluations.

\end{document}